%% file: main.tex
\pdfoutput=1

\documentclass[11pt]{article}
\usepackage{xcolor}
\usepackage{algorithm}
\usepackage{xcolor}
\usepackage[colorlinks=true, urlcolor=blue]{hyperref}
\usepackage{multirow}
\usepackage{booktabs}
\usepackage{verbatim}
 \usepackage{fancyvrb}
\usepackage{algorithm}
\usepackage{algorithmic}
\usepackage{boxedminipage}
\usepackage{listings}
\usepackage[final]{acl}
\usepackage{orcidlink}

\usepackage{times}
\usepackage{latexsym}

\usepackage[T1]{fontenc}

\usepackage[utf8]{inputenc}

\usepackage{microtype}
\usepackage{amsmath}

\usepackage{inconsolata}

\usepackage{graphicx}
\usepackage{float}
\usepackage{placeins}

\title{ttda704 at SemEval-2026 Task 6: Structured Chain-of-Thought Prompting for Political Evasion Detection}

\author{
 \textbf{Tai Tran Tan\textsuperscript{1,2}\textsuperscript{*}\orcidlink{0009-0005-4008-6317}},
 \textbf{An Dinh Thien\textsuperscript{1,2}\textsuperscript{*}\orcidlink{0009-0003-9909-8852}}
\\
 \textsuperscript{1}University of Information Technology, Ho Chi Minh City, Vietnam\\
 \textsuperscript{2}Vietnam National University, Ho Chi Minh City, Vietnam
\\
   \{22521287, 22520010\}@gm.uit.edu.vn
}

\begin{document}
\maketitle
\begingroup
\renewcommand{\thefootnote}{*}
\footnotetext{Equal contributions.}
\endgroup
\setcounter{footnote}{0}

\input{sec/0_abstract}    
\input{sec/1_introduction}
\input{sec/2_related-work}
\input{sec/3_shared_task_description}
\input{sec/4_method}
\input{sec/5_results}
\input{sec/6_conclusion}
\bibliography{custom}

\input{sec/7_appendix}

\end{document}

%% file: sec/0_abstract.tex
\begin{abstract}
    This paper describes our system\footnote{Code and prompts are available at \url{https://github.com/taitran501/SemEval-2026-Task6}} for SemEval-2026 Task 6, which addresses the classification of political evasion strategies in English question--answer pairs extracted from U.S.\ presidential interviews. We systematically compare two distinct paradigms: (1)~Parameter-Efficient Fine-Tuning of Qwen3 models (4B--32B) using QLoRA, enhanced with tiered upsampling and weighted cross-entropy loss to address severe class imbalance, and (2)~structured Chain-of-Thought (CoT) prompting of reasoning-capable API models, namely DeepSeek-V3.2 and Grok-4-Fast. Our evaluation demonstrates that structured CoT prompting of reasoning-enabled models substantially outperforms our baseline parameter-efficient fine-tuning implementation in absolute Macro~F1. Our best system, Grok-4-Fast with extended reasoning and few-shot hierarchical CoT prompting, achieves a Macro~F1 of 0.5147 on Subtask~2 (9-class evasion) and 0.7979 on Subtask~1 (3-class clarity), ranking 8/33 on Subtask~2 and 13/41 on Subtask~1 on the official leaderboard. Furthermore, our ablation studies reveal key insights into effective prompt design for evasion detection: presenting labels within a hierarchical taxonomy helps structure model reasoning while few-shot exemplars provide task calibration; however, the strongest prompt variants are not statistically distinguishable in Macro~F1, and explicitly enabling extended reasoning modes yields substantial performance gains by facilitating the multi-step pragmatic analysis required to detect evasive intent.
    \end{abstract}

%% file: sec/1_introduction.tex
\section{Introduction}
\label{sec:intro}

Political discourse is strategically ambiguous. In televised interviews, politicians routinely employ evasion techniques such as topic shifts, deflections, and refusals that undermine democratic transparency. Prior work reports that politicians provide clear answers to only 39--46\% of questions, compared to 70--89\% for non-politicians \citep{bull2003microanalysis}. Automatically detecting these strategies is challenging because evasion often relies on pragmatic inference rather than surface-level cues.

SemEval-2026 Task~6 \citep{thomas-etal-2024-never}, \textit{CLARITY: Unmasking Political Question Evasions}, focuses on classifying semantic evasion in political question-answer (QA) pairs. The task poses two evaluation challenges: \textbf{Subtask 1 (Clarity)} categorizes responses into three coarse tiers (\emph{Clear Reply}, \emph{Ambivalent}, \emph{Clear Non-Reply}), while \textbf{Subtask 2 (Evasion)} demands fine-grained classification into nine specific evasion strategies (e.g., \emph{Dodging}, \emph{Deflection}, \emph{Implicit}). 

Standard text classifiers handle overt evasion but struggle with subtle, ambivalent replies because they lack explicit pragmatic reasoning. Separating a \emph{General} platitude from an \emph{Implicit} answer, or a tangential \emph{Deflection} from full \emph{Dodging}, requires context-sensitive inference beyond surface matching.

\begin{figure}[t]
    \centering
    \includegraphics[width=\linewidth]{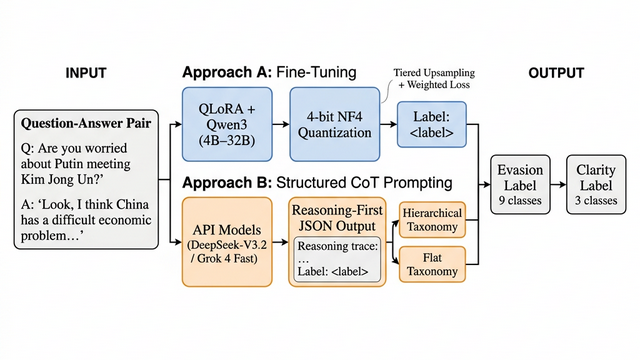}
    \caption{Overview of our two pipelines: Approach~A (QLoRA fine-tuning) and Approach~B (structured CoT prompting).}
    \label{fig:approach-overview}
\end{figure}

We address this by systematically studying political evasion classification via two paradigms: PEFT of open-weight models (Qwen3 4B--32B) and structured CoT prompting with reasoning-focused APIs (DeepSeek-V3.2, Grok-4-Fast). To mitigate the difficulty of the flat 9-class taxonomy, we introduce hierarchical prompts that evaluate ``Directness'' and ``Topic Fidelity'' before the final label (Figure~\ref{fig:approach-overview}).

Our main contributions are as follows:
\begin{enumerate}
    \item We identify that the lack of intermediate pragmatic reasoning is a core bottleneck for classifying ambivalent evasion. To overcome this, we introduce step-by-step CoT prompting templates that explicitly condition models to evaluate linguistic directness and topic alignment.
    \item We show that restructuring a flat class taxonomy into a conceptual hierarchy provides a useful reasoning scaffold for distinguishing closely related evasion boundaries, helping capable models organize fine-grained labels and produce clearer reasoning traces.
    \item We analyze model errors and show that CoT reasoning can over-interpret vague political language, especially around \emph{Dodging}, \emph{Deflection}, \emph{General}, and \emph{Implicit}.
\end{enumerate}

%% file: sec/2_related-work.tex
\section{Related Work}

Political evasion has long been studied in discourse and political communication, where prior work documents how public figures avoid, partially answer, or reframe questions in interviews \citep{bull2003microanalysis,bull1993hownotanswer,clayman2001answers,rasiah2010frameworkevasion}. Recent NLP work operationalizes this phenomenon through response-clarity and evasion taxonomies, including the QEvasion dataset and the CLARITY shared task \citep{thomas-etal-2024-never,ferracane2021didanswer}.

Our work relates to two lines of NLP research. First, hierarchical label information can help models reason over related political labels \citep{dayanik2022improving}. Second, chain-of-thought and structured prompting can elicit multi-step reasoning in LLMs \citep{wei2022chain,wang2023plan}, while PEFT methods such as LoRA and QLoRA enable efficient adaptation of open-weight models \citep{hu2022lora,dettmers2023qlora}. We compare these two paradigms under a controlled shared-task setting for political evasion detection.

%% file: sec/3_shared_task_description.tex
\section{Task Description}
\label{sec:task}

\subsection{Task Definition}
SemEval-2026 Task~6 \citep{thomas-etal-2024-never}, \textit{CLARITY: Unmasking Political Question Evasions}, frames the detection of political evasion as a classification task over question--answer (QA) pairs drawn from U.S.\ presidential interviews spanning 2006--2023. Multi-part interview questions are first decomposed into singular sub-questions, so that each annotation precisely captures how well a single specific inquiry is addressed. Systems are evaluated on two nested subtasks:

\begin{itemize}
    \item \textbf{Subtask 1 (Clarity):} Classify the respondent's answer into one of three coarse clarity tiers: \textit{Clear Reply}, \textit{Ambivalent}, or \textit{Clear Non-Reply}.
    \item \textbf{Subtask 2 (Evasion):} Classify the answer into one of nine fine-grained evasion strategies. The clarity label is deterministically derived from the evasion label via a fixed mapping.
\end{itemize}

The nine evasion labels and their parent clarity categories are summarized in Table~\ref{tab:taxonomy}.

\begin{table}[h]
\centering
\small
{\setlength{\tabcolsep}{4pt}
\begin{tabular}{l l}
\toprule
\textbf{Clarity (Subtask 1)} & \textbf{Evasion (Subtask 2)} \\
\midrule
Clear Reply      & Explicit \\
\midrule
Ambivalent       & Implicit \\
                 & General \\
                 & Dodging \\
                 & Deflection \\
                 & Partial/half-answer \\
\midrule
Clear Non-Reply  & Declining to answer \\
                 & Claims ignorance \\
                 & Clarification \\
\bottomrule
\end{tabular}
}
\caption{The two-level taxonomy of response clarity. Each evasion label (Subtask 2) maps to exactly one clarity category (Subtask 1).}
\label{tab:taxonomy}
\end{table}

\subsection{Evaluation}
Both subtasks are evaluated using macro-averaged F1-score, ensuring equal weight across all classes regardless of their frequency. Evaluation is conducted on the official test set released by the organizers, as well as a held-out private evaluation set used for the final competition leaderboard.

\subsection{Dataset}
The dataset \citep{thomas-etal-2024-never} contains 3,448 training samples and 308 test samples. The training set is heavily imbalanced: \textit{Explicit} (30.5\%) and \textit{Dodging} (20.5\%) are the dominant classes, while \textit{Partial/half-answer} (2.3\%) and \textit{Clarification} (2.7\%) are severely underrepresented. Each instance provides three text fields: the full original interview question (\texttt{interview\_question}), a decomposed sub-question (\texttt{question}), and the respondent's full answer (\texttt{interview\_answer}).

%% file: sec/4_method.tex
\section{Methodology}
\label{sec:method}

\subsection{System Overview}
We explore two complementary approaches: \textbf{Approach~A} (Parameter-Efficient Fine-Tuning of local models) and \textbf{Approach~B} (structured Chain-of-Thought prompting of API models). Both approaches emphasize explicit reasoning as a central mechanism for evasion detection \citep{wei2022chain,wang2023plan}. Figure~\ref{fig:approach-overview} summarizes the two pipelines.

\subsection{Approach A: Parameter-Efficient Fine-Tuning}
We fine-tune the Qwen3 family \citep{yang2025qwen3technicalreport} using QLoRA with 4-bit NF4 quantization \citep{dettmers2023qlora}. To counteract severe class imbalance, we apply 2$\times$ upsampling to the three rarest evasion classes and apply inverse-frequency weighting (clipped at 5.0) to the cross-entropy loss. We evaluate four variants: Qwen3-4B, Qwen3-4B-Instruct-2507, Qwen3-14B, and Qwen3-32B.

\textbf{Training configuration.} 
Models are trained for a single epoch using Unsloth \citep{han2024unsloth} with learning rate $2 \times 10^{-5}$ and a cosine scheduler. LoRA rank is set to 32 (alpha 64) across all linear projection layers. We proactively limit training to one epoch to minimize the risk of overfitting on the relatively small training set. Full per-model hyperparameters are detailed in Appendix~\ref{sec:appendix-setup}.

\subsection{Approach B: Structured Chain-of-Thought Prompting}
For API models, we enforce a reasoning-first output format consistent with prior CoT work \citep{wei2022chain}. We evaluate two state-of-the-art architectures: DeepSeek-V3.2 (\texttt{deepseek-chat} vs.\ \texttt{deepseek-reasoner}) and Grok-4-Fast (\texttt{grok-4-fast} vs.\ \texttt{grok-4-fast-reasoning}). For both providers, we compare standard inference against extended reasoning modes that generate an internal chain-of-thought before the final answer. All models strictly output JSON containing a \texttt{reasoning} field followed by the \texttt{evasion\_label} \citep{li2025reinforcement}.

\subsection{Prompt Design}
We provide the full \texttt{interview\_question} as context and the decomposed \texttt{question} as the target, following \citet{thomas-etal-2024-never}. Prompts vary by in-context learning (zero-shot vs.\ few-shot: 9 exemplars for API CoT, 6 for fine-tuned standard) and taxonomy structure (hierarchical vs.\ flat).

Fine-tuned models output either standard text (`\textit{Label: <label>}') or structured JSON (CoT). Together with the prompt axes, this yields four configurations per model. Prompt templates are detailed in Appendix~\ref{sec:appendix-prompts}.

%% file: sec/5_results.tex
\section{Results and Discussion}
\label{sec:results}

\subsection{Experimental Setup}
We evaluate all models on the 308-sample official test split. We report results using majority-vote gold labels from three independent annotators. Macro~F1 is the primary metric for both subtasks. Subtask~1 F1 is derived by deterministically mapping predicted evasion labels to their parent clarity categories. For fine-tuned models with CoT outputs, labels are extracted using a robust three-stage pipeline of regex matching and string fallback (Appendix~\ref{sec:appendix-setup}).

\subsection{Overview of Results}

Table~\ref{tab:overview} summarizes the key systems; the full 32-configuration matrix is provided in Appendix~\ref{sec:appendix-full-results}.

\begin{table}[t]
\centering
\footnotesize
\setlength{\tabcolsep}{3pt}
\begin{tabular}{p{0.50\columnwidth}cc}
\toprule
\textbf{System} & \textbf{S2 F1} & \textbf{S1 F1} \\
\midrule
Best Qwen3 PEFT & 0.3630 & 0.5654 \\
DeepSeek Rsn. Few-shot Hier. & 0.5115 & 0.7735 \\
Grok Rsn. Zero-shot Flat & 0.5126 & 0.7821 \\
Grok Rsn. Few-shot Hier. & \textbf{0.5147} & \textbf{0.7979} \\
Official submission$^\dagger$ & 0.5600 & 0.7900 \\
Organizer revised baseline$^\dagger$ & 0.5700 & 0.8200 \\
\bottomrule
\end{tabular}
\caption{Key results. S2 and S1 denote Macro~F1 for Subtask~2 (evasion) and Subtask~1 (clarity), respectively. The top four rows are results on the 308-sample official test split with majority-vote labels. Rows marked with $^\dagger$ are blind/private leaderboard scores and are shown only for context. The full 32-configuration comparison is in Appendix~\ref{sec:appendix-full-results}.}
\label{tab:overview}
\end{table}

The best overall configuration on the 308-sample official test split is \textbf{Grok-4-Fast Reasoning, Few-shot CoT (hier.)}, achieving Macro~F1 of \textbf{0.5147} on Subtask~2 and \textbf{0.7979} on Subtask~1. The best fine-tuned model (Qwen3-32B, Few-shot CoT) reaches 0.3630 on Subtask~2, so structured CoT prompting with reasoning-capable API models substantially outperforms our \emph{baseline} PEFT setup. We interpret this comparison conservatively because it is confounded by model scale, disabled Qwen thinking mode, a single fine-tuning epoch, and different exemplar counts.

\paragraph{Official leaderboard.}
Our best submission (\textit{ttda704}) ranks 8/33 on Subtask~2 (Evasion F1\,=\,0.56) and 13/41 on Subtask~1 (Clarity F1\,=\,0.79) on the official SemEval-2026 Task~6 leaderboard. The revised organizer baseline reported in the camera-ready task paper \citep{thomas2026semeval2026task6clarity} is 0.57 on Subtask~2 and 0.82 on Subtask~1.

\paragraph{Subtask 1 behavior.}
Subtask~1 scores are consistently higher because the coarse clarity taxonomy collapses several difficult ambivalent evasion labels into a single parent class. Errors such as \textit{General} vs.\ \textit{Implicit} or \textit{Dodging} vs.\ \textit{Deflection} are severe under the nine-way Subtask~2 metric, but often remain within the \emph{Ambivalent} category and therefore do not hurt Subtask~1. This explains why models achieve relatively strong clarity F1 while still struggling with fine-grained evasion distinctions.

\subsection{Ablation Analysis}

\textbf{Reasoning mode.} Enabling extended reasoning consistently improves both API model families, with the largest gains under few-shot CoT. The benefit is especially clear for Grok, where reasoning plus zero-shot flat prompting already reaches 0.5126 Macro~F1, nearly matching the strongest few-shot configuration.

\paragraph{Prompt structure and exemplars.}
The hierarchical taxonomy is best interpreted as a reasoning scaffold rather than a universally superior prompt format. It encourages models to first decide coarse clarity before selecting a fine-grained evasion label, which appears most useful for reasoning-capable models. Few-shot exemplars further calibrate ambiguous boundaries such as \textit{General} vs.\ \textit{Implicit} and \textit{Dodging} vs.\ \textit{Deflection}. However, the effect is not uniform: Grok-4-Fast Reasoning with zero-shot flat prompting reaches 0.5126 Macro~F1, nearly matching the best few-shot hierarchical configuration at 0.5147.

\paragraph{Statistical significance.}
We assess the top configurations using approximate randomization tests \citep{yeh2000more} ($R=10{,}000$), McNemar's test \citep{mcnemar1947note}, and paired bootstrap confidence intervals ($B=10{,}000$). The top three systems have heavily overlapping 95\% confidence intervals: Grok Few-shot Hier. $[0.443,0.565]$, Grok Zero-shot Flat $[0.436,0.567]$, and DeepSeek Few-shot Hier. $[0.428,0.569]$. Approximate randomization tests show no significant Macro~F1 differences ($p>0.19$), so the best configuration should be interpreted as the highest point estimate rather than a statistically superior system. McNemar's test nevertheless shows significant sample-level disagreement between Grok and DeepSeek few-shot hierarchical systems ($p=0.015$), suggesting complementary error patterns.

\textbf{Fine-tuned models.} For local Qwen models, CoT prompting is not uniformly helpful, especially at 4B scale. The combination of simplified reasoning targets and output-format fragility means these results should be interpreted as a PEFT baseline rather than an upper bound.

\subsection{Error Analysis}
\label{sec:error}

Table~\ref{tab:per-class} reports per-class performance for the best system. The full confusion matrix and qualitative case studies are provided in Appendix~\ref{sec:appendix-cm}.

\begin{table}[t]
\centering
\small
{\setlength{\tabcolsep}{3pt}
\begin{tabular}{lccc}
\toprule
Evasion Label & Prec & Rec & F1 \\
\midrule
Claims ignorance      & 0.7778 & 0.8750 & 0.8235 \\
Clarification         & 1.0000 & 1.0000 & 1.0000 \\
Declining to answer   & 0.6923 & 0.7500 & 0.7200 \\
Deflection            & 0.2200 & 0.5238 & 0.3099 \\
Dodging               & 0.8333 & 0.1852 & 0.3030 \\
Explicit              & 0.6636 & 0.8022 & 0.7264 \\
General               & 0.3571 & 0.2885 & 0.3191 \\
Implicit              & 0.3103 & 0.3000 & 0.3051 \\
Partial/half-answer   & 0.1000 & 0.1667 & 0.1250 \\
\bottomrule
\end{tabular}
}
\caption{Per-class Subtask~2 performance for the best system (Grok-4-Fast Reasoning, Few-shot CoT) on the test set (N\,=\,308).}
\label{tab:per-class}
\end{table}

Performance drops sharply on pragmatically subtle categories. We observe three recurrent failure modes:

\paragraph{Dodging vs.\ other ambivalent forms.}
\textit{Dodging} has critically low recall (0.1852), indicating that the model rarely identifies complete topic abandonment. When a true \textit{Dodging} response is misclassified, the model predicts \textit{Implicit} 26\%, \textit{General} 22\%, and \textit{Deflection} 20\% of the time. A concrete mitigation is a second-stage topic-abandonment verifier that checks whether the answer preserves the core entity and requested information from the target question.

\paragraph{General vs.\ Implicit vs.\ Deflection.}
The model exhibits high mutual confusion across these ambivalent categories. Of true \textit{General} errors, 29\% are predicted as \textit{Implicit} and 27\% as \textit{Deflection}, suggesting that step-by-step reasoning can over-extract specific intent from vague political language.

\paragraph{Over-prediction of Explicit.}
The model frequently defaults to the majority class. Notably, 42\% of true \textit{Implicit} responses are misclassified as \textit{Explicit}, suggesting that topical relevance is often treated as sufficient evidence of directness. Appendix~\ref{sec:appendix-cs3} also shows that some apparent errors reflect genuine annotator disagreement rather than clear model failure.

\subsection{Thinking Mode Follow-up (Qwen3-32B)}
\label{sec:thinking-followup}

To address the confound that Qwen3's native thinking mode was disabled in the main PEFT experiments, we ran a diagnostic follow-up on a Qwen3-32B checkpoint. Because thinking-mode inference was substantially slower (approximately 100 minutes for 100 samples in our setup), we evaluated it on stratified samples of 100 instances rather than rerunning the full test set or full configuration grid. Enabling thinking improved Evasion F1 from 0.0734 to 0.2362 on the official-test sample setting and reduced hard fallback from 37.99\% to 0.00\%. On the blind sample, thinking reached 0.3700 Evasion F1 with a 1.00\% hard-fallback rate. These results suggest that the original PEFT baseline is conservative, although this experiment is diagnostic rather than fully comparable to the main benchmark. Full details are reported in Appendix~\ref{sec:appendix-thinking} and Appendix~\ref{sec:appendix-fallback}.

%% file: sec/6_conclusion.tex
\section{Conclusion}
\label{sec:conclusion}

We compared QLoRA-based PEFT and structured CoT prompting for SemEval-2026 Task~6. Reasoning-capable API models substantially outperform our baseline PEFT setup in absolute Macro~F1, although the comparison is confounded by model scale, native reasoning support, and exemplar count. Our best configuration, Grok-4-Fast Reasoning with few-shot hierarchical CoT, achieves the highest point estimate (0.5147 on Subtask~2 and 0.7979 on Subtask~1), but is not statistically superior to the strongest zero-shot flat variant. Overall, our findings suggest that hierarchical prompts are useful as reasoning scaffolds, few-shot examples act as task calibration, and the hardest remaining errors arise from pragmatic ambiguity around ambivalent evasion labels.

\subsection*{Limitations}
Our PEFT results should be interpreted as a baseline rather than an upper bound: models were trained for one epoch without extensive hyperparameter search, and Qwen3 thinking mode was disabled in the main experiments. Our few-shot examples were manually curated and not systematically varied, which may introduce exemplar and speaker-style bias. Future work should evaluate exemplar sensitivity by sampling multiple few-shot sets across different presidents and by retrieving semantically similar exemplars at inference time; retrieval-based few-shot prompting may reduce speaker-style bias and improve coverage of rare evasion labels. We report blind/private leaderboard scores only for the submitted system rather than rerunning all 32 configurations, because full blind-set reruns would require substantial additional API calls and local inference time; controlled ablations are therefore conducted on the 308-sample official test split. Finally, the best result relies on Grok-4-Fast, a proprietary API model; the strongest open-weight reproducible configuration is DeepSeek-V3.2 Reasoning Few-shot CoT (hier.) at Macro~F1\,=\,0.5115 on Subtask~2.

%% file: sec/7_appendix.tex
\appendix

\section{Detailed Experimental Setup}
\label{sec:appendix-setup}

\subsection{Fine-Tuning Hyperparameters}

Table~\ref{tab:hyperparams} reports the full hyperparameter configuration for each fine-tuned model. The specific model checkpoints used are: \texttt{unsloth/\allowbreak Qwen3-4B} (4B base), \texttt{unsloth/\allowbreak Qwen3-4B-Instruct-2507-\allowbreak unsloth-\allowbreak bnb-\allowbreak 4bit} (4B-Instruct), \texttt{unsloth/\allowbreak Qwen3-14B-\allowbreak unsloth-\allowbreak bnb-\allowbreak 4bit} (14B), and \texttt{unsloth/\allowbreak Qwen3-32B-\allowbreak unsloth-\allowbreak bnb-\allowbreak 4bit} (32B). All models share common LoRA parameters (rank 32, alpha 64, dropout 0.0) applied to all linear layers (\texttt{q\_proj}, \texttt{k\_proj}, \texttt{v\_proj}, \texttt{o\_proj}, \texttt{gate\_proj}, \texttt{up\_proj}, \texttt{down\_proj}). Training uses AdamW 8-bit optimizer with cosine learning rate scheduling. Qwen3 thinking mode is disabled (\texttt{enable\_thinking=False}) during both training and inference for all models.

\begin{table}[h]
\centering
\small
{\setlength{\tabcolsep}{3pt}
\begin{tabular}{lcccc}
\toprule
Parameter & 4B & 4B-Inst & 14B & 32B \\
\midrule
Quantization         & \multicolumn{4}{c}{NF4 (4-bit)} \\
LoRA rank / alpha    & \multicolumn{4}{c}{32 / 64} \\
LoRA dropout         & \multicolumn{4}{c}{0.0} \\
Learning rate        & \multicolumn{4}{c}{$2 \times 10^{-5}$} \\
LR scheduler         & \multicolumn{4}{c}{Cosine} \\
Warmup ratio         & \multicolumn{4}{c}{0.05} \\
Epochs               & \multicolumn{4}{c}{1} \\
Optimizer            & \multicolumn{4}{c}{AdamW 8-bit} \\
Max class weight     & \multicolumn{4}{c}{5.0} \\
\midrule
Batch / GPU          & 16  & 16  & 2   & 2   \\
Grad.\ accum.        & 2   & 2   & 8   & 8   \\
Eff.\ batch size     & 32  & 32  & 16  & 16  \\
Weight decay         & 0.01 & 0.01 & 0.05 & 0.05 \\
Max grad norm        & 1.0  & 1.0  & 0.5  & 0.5  \\
Max seq.\ length     & 8092 & 8092 & 7000 & 7000 \\
Max new tokens       & 1024 & 1024 & 2048 & 2048 \\
\bottomrule
\end{tabular}
}
\caption{Full hyperparameter configuration for fine-tuned models. All models are trained with Unsloth for memory-efficient QLoRA.}
\label{tab:hyperparams}
\end{table}

\subsection{Dataset Statistics and Inter-Annotator Agreement}

\begin{figure}[h]
\centering
\includegraphics[width=\linewidth]{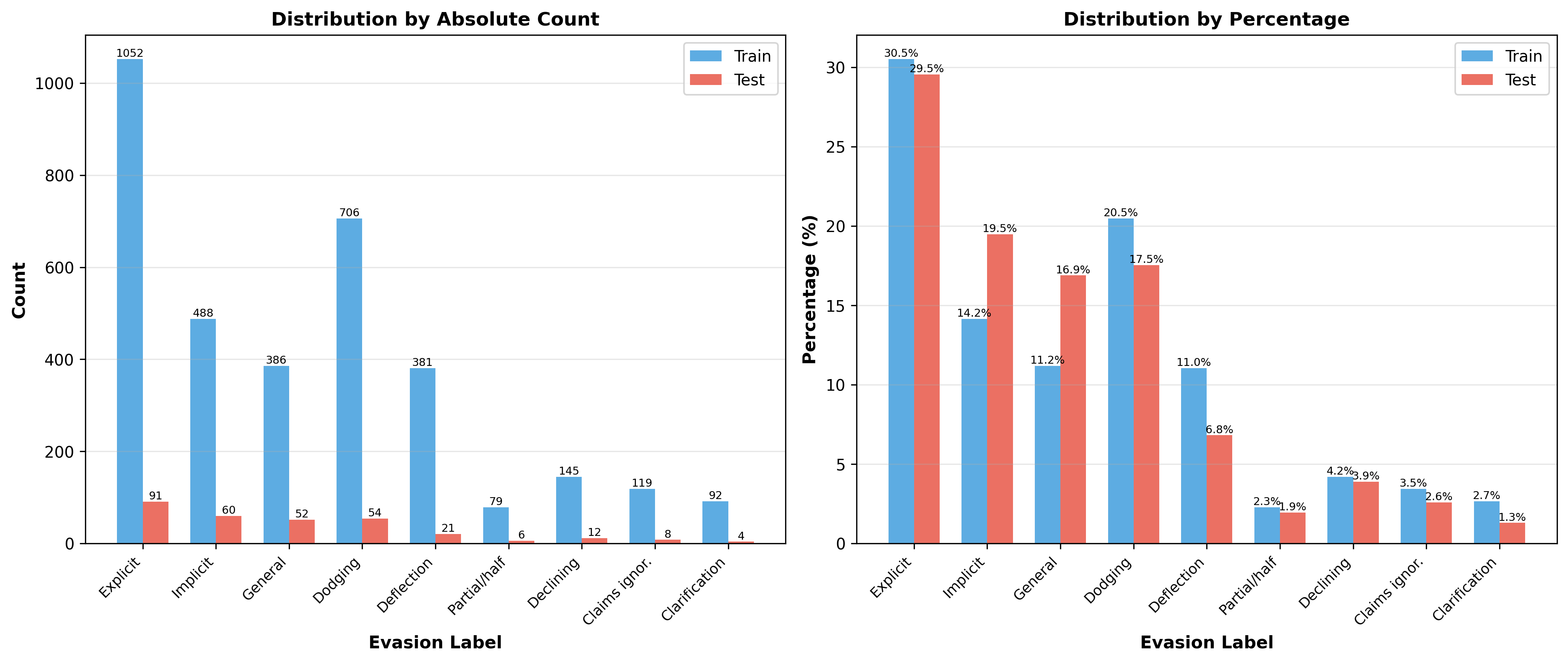}
\caption{Visualization of label distribution across training and test sets, showing severe class imbalance particularly for the three rarest categories.}
\label{fig:data-dist}
\end{figure}

Table~\ref{tab:data-stats} presents the complete distribution of evasion labels across training and test sets, revealing severe class imbalance with a 13.32:1 ratio between the most frequent (\textit{Explicit}, 1052 samples) and least frequent (\textit{Partial/half-answer}, 79 samples) classes in the training set. Figure~\ref{fig:data-dist} visualizes this distribution, highlighting that the three rarest classes (\textit{Clarification}, \textit{Claims ignorance}, \textit{Partial/half-answer}) collectively represent only 8.41\% of the training data.

\begin{table}[h]
\centering
\footnotesize
{\setlength{\tabcolsep}{2.5pt}
\begin{tabular}{lrrrr}
\toprule
\textbf{Label} & \textbf{Train} & \textbf{Train\%} & \textbf{Test} & \textbf{Test\%} \\
\midrule
Explicit              & 1052 & 30.51 & 91  & 29.55 \\
Implicit              & 488  & 14.15 & 60  & 19.48 \\
General               & 386  & 11.19 & 52  & 16.88 \\
Dodging               & 706  & 20.48 & 54  & 17.53 \\
Deflection            & 381  & 11.05 & 21  & 6.82  \\
Partial/half-ans.     & 79   & 2.29  & 6   & 1.95  \\
Declining to ans.     & 145  & 4.21  & 12  & 3.90  \\
Claims ignorance      & 119  & 3.45  & 8   & 2.60  \\
Clarification         & 92   & 2.67  & 4   & 1.30  \\
\midrule
\textbf{Total}        & 3448 & 100.00 & 308 & 100.00 \\
\bottomrule
\end{tabular}
}
\caption{Distribution of evasion labels in training and test sets. The class imbalance ratio is 13.32:1 (Explicit to Partial/half-answer).}
\label{tab:data-stats}
\end{table}

\paragraph{Inter-Annotator Agreement.}
The dataset includes annotations from three independent annotators per instance. For the test set (N\,=\,308), we observe full agreement (3/3 annotators) in 40.58\% of cases, partial agreement (2/3 annotators) in 48.70\%, and complete disagreement (all three different) in 10.71\%. We calculate Fleiss' Kappa at 0.4723, indicating \textit{moderate agreement}. This level of agreement is expected given the inherently subjective nature of pragmatic interpretation - human judges may legitimately focus on different aspects of the same response when applying the taxonomy.

The most common disagreement patterns reveal systematic ambiguities in the taxonomy boundaries: \textit{General} vs. \textit{Implicit} (24 cases), \textit{Explicit} vs. \textit{Implicit} (23 cases), and \textit{Deflection} vs. \textit{Dodging} (16 cases). These patterns directly correspond to the confusion clusters observed in our model predictions (Section~\ref{sec:error}), suggesting that even human annotators struggle with the same pragmatic boundaries that confound automated systems. Overall, 183 instances (4.87\% of the full dataset) exhibit annotator disagreement, representing cases where political discourse interpretation is genuinely ambiguous rather than algorithmically deterministic.

\subsection{Class Imbalance Handling}
We address severe class imbalance via two complementary techniques:
\begin{enumerate}
    \item \textbf{Tiered upsampling (2$\times$)} of the three rarest classes: \textit{Clarification} (92 $\rightarrow$ 184), \textit{Claims ignorance} (119 $\rightarrow$ 238), and \textit{Partial/half-answer} (79 $\rightarrow$ 158).
    \item \textbf{Inverse-frequency class weights} computed as $w_c = \frac{N}{C \cdot n_c}$ (where $N$ = total samples, $C$ = number of classes, $n_c$ = count of class $c$), clipped at 5.0 to avoid numerical instability.
\end{enumerate}

\subsection{Label Extraction Pipeline}
For fine-tuned models generating CoT outputs, we extract evasion labels with a three-stage pipeline:
\begin{enumerate}
    \item Regex match for \texttt{"evasion\_label": "<label>"} in JSON output.
    \item Fallback: extract text after \texttt{Label:} prefix.
    \item Final fallback: string-match any valid label in the full output. If no label is found, default to \textit{Explicit} (most frequent class).
\end{enumerate}

\noindent This fallback is most frequently triggered for Qwen3-4B with CoT output format, where JSON parsing failures occur due to the model's limited instruction-following at small scale. The fallback may inflate Explicit-class precision and recall for these configurations.

\subsection{Hard-Fallback Statistics}
\label{sec:appendix-fallback}

Table~\ref{tab:fallback} reports the hard-fallback rate for the Qwen3-32B model under different inference configurations. A hard fallback (defaulting to \textit{Explicit}) indicates a complete output format failure rather than a genuine prediction, and its rate therefore directly reflects the reliability of the extraction pipeline for a given configuration.

\begin{table*}[t]
\centering
\small
{\setlength{\tabcolsep}{6pt}
\begin{tabular*}{\textwidth}{@{\extracolsep{\fill}} l l c c c}
\toprule
\textbf{Model} & \textbf{Inference Mode} & \textbf{Split} & \textbf{N} & \textbf{Hard Fallback} \\
\midrule
Qwen3-32B & Non-thinking & Official Test & 308 & 37.99\% (117/308) \\
Qwen3-32B & Thinking     & Official Test & 100 & \textbf{0.00\%} (0/100) \\
Qwen3-32B & Thinking     & Blind Set     & 100 & \textbf{1.00\%} (1/100) \\
\bottomrule
\end{tabular*}
}
\caption{Hard-fallback rates for Qwen3-32B. The high rate in non-thinking mode (37.99\%) indicates systematic output format failure; enabling thinking inference reduces this to near zero, demonstrating that the reasoning mode substantially stabilises output formatting in addition to improving classification quality.}
\label{tab:fallback}
\end{table*}

The near-complete elimination of hard fallbacks when thinking is enabled suggests two effects: (1) the model's extended reasoning process produces more structured, parseable outputs; and (2) the non-thinking hard-fallback rate artificially inflates \textit{Explicit}-class predictions in the main results table, making the non-thinking baseline appear worse than a configuration with stable output formatting.

We report the Qwen3-32B diagnostic setting because it directly affects the thinking-mode follow-up in Section~\ref{sec:results}. A full per-configuration fallback audit would require reprocessing all raw generations and is outside the scope of this camera-ready revision; however, structured CoT outputs are the most affected by complete parsing failure, whereas standard label-only prompting is less prone to hard fallback.

\subsection{API Model Configuration}
{\sloppy
\textbf{DeepSeek-V3.2} is accessed via the DeepSeek API. For non-thinking inference, we use the model \texttt{deepseek-chat} with \texttt{temperature=0}. For thinking inference, we use \texttt{deepseek-reasoner}. This mode produces an internal chain-of-thought in the \texttt{reasoning\_content} field before the final answer. In both settings, we request JSON output with \texttt{response\_format} set to \texttt{\{"type":"json\_object"\}}.

\textbf{Grok-4-Fast} is accessed via the xAI batch API. Requests are chunked into batches of up to 50{,}000 items, uploaded as JSONL, and retrieved after completion. The non-reasoning configuration uses \texttt{grok-4-fast}; the reasoning configuration uses \texttt{grok-4-fast-reasoning} with extended thinking tokens. All requests use \texttt{temperature=0}.
}

\section{Additional Results}
\label{sec:appendix-additional-results}

\subsection{Full Results Matrix}
\label{sec:appendix-full-results}

Table~\ref{tab:overview-full} reports the complete 32-configuration comparison on the official test set.

\begin{table*}[t]
\centering
\small
\setlength{\tabcolsep}{4pt}
\begin{tabular}{l l l c c}
\toprule
\textbf{Model} & \textbf{Mode} & \textbf{Prompt strategy} & \textbf{Evasion F1 (Subtask 2)} & \textbf{Clarity F1 (Subtask 1)} \\
\midrule
\multicolumn{5}{l}{\textit{Approach A: Fine-tuned Models}} \\
\midrule
Qwen3-4B       & Base           & Zero-shot              & 0.2302 & 0.4459 \\
Qwen3-4B       & Base           & Few-shot               & 0.1400 & 0.4203 \\
Qwen3-4B       & Base           & Zero-shot CoT          & 0.0902 & 0.2455 \\
Qwen3-4B       & Base           & Few-shot CoT           & \underline{0.2455} & \underline{0.5682} \\
\addlinespace
Qwen3-4B       & Instruct       & Zero-shot              & \underline{0.3384} & 0.5307 \\
Qwen3-4B       & Instruct       & Few-shot               & 0.2327 & \underline{0.5338} \\
Qwen3-4B       & Instruct       & Zero-shot CoT          & 0.1064 & 0.2249 \\
Qwen3-4B       & Instruct       & Few-shot CoT           & 0.1738 & 0.4590 \\
\addlinespace
Qwen3-14B      & Base           & Zero-shot              & \underline{0.3201} & \underline{0.5671} \\
Qwen3-14B      & Base           & Few-shot               & 0.3150 & 0.5628 \\
Qwen3-14B      & Base           & Zero-shot CoT          & 0.3022 & 0.5073 \\
Qwen3-14B      & Base           & Few-shot CoT           & 0.3122 & 0.5047 \\
\addlinespace
Qwen3-32B      & Base           & Zero-shot              & 0.3220 & 0.5879 \\
Qwen3-32B      & Base           & Few-shot               & 0.3215 & 0.5920 \\
Qwen3-32B      & Base           & Zero-shot CoT          & 0.3288 & \underline{0.6126} \\
Qwen3-32B      & Base           & Few-shot CoT           & \underline{0.3630} & 0.5654 \\
\midrule
\multicolumn{5}{l}{\textit{Approach B: API Models}} \\
\midrule
DeepSeek-V3.2  & Non-reasoning  & Zero-shot CoT (hier.)  & 0.2646 & 0.4998 \\
DeepSeek-V3.2  & Non-reasoning  & Zero-shot CoT (flat)   & 0.3588 & 0.5902 \\
DeepSeek-V3.2  & Non-reasoning  & Few-shot CoT (hier.)   & \underline{0.4578} & \underline{0.7605} \\
DeepSeek-V3.2  & Non-reasoning  & Few-shot CoT (flat)    & 0.4092 & 0.6968 \\
\addlinespace
DeepSeek-V3.2  & Reasoning      & Zero-shot CoT (hier.)  & 0.3085 & 0.6030 \\
DeepSeek-V3.2  & Reasoning      & Zero-shot CoT (flat)   & 0.3795 & 0.6275 \\
DeepSeek-V3.2  & Reasoning      & Few-shot CoT (hier.)   & \underline{0.5115} & \underline{0.7735} \\
DeepSeek-V3.2  & Reasoning      & Few-shot CoT (flat)    & 0.4714 & 0.7369 \\
\addlinespace
Grok-4-Fast    & Non-reasoning  & Zero-shot CoT (hier.)  & 0.2636 & 0.4909 \\
Grok-4-Fast    & Non-reasoning  & Zero-shot CoT (flat)   & 0.4075 & 0.6746 \\
Grok-4-Fast    & Non-reasoning  & Few-shot CoT (hier.)   & 0.4071 & \underline{0.7088} \\
Grok-4-Fast    & Non-reasoning  & Few-shot CoT (flat)    & \underline{0.4157} & 0.6742 \\
\addlinespace
Grok-4-Fast    & Reasoning      & Zero-shot CoT (hier.)  & 0.3015 & 0.6470 \\
Grok-4-Fast    & Reasoning      & Zero-shot CoT (flat)   & 0.5126 & 0.7821 \\
Grok-4-Fast    & Reasoning      & Few-shot CoT (hier.)   & \textbf{0.5147} & \textbf{0.7979} \\
Grok-4-Fast    & Reasoning      & Few-shot CoT (flat)    & 0.5043 & 0.7565 \\
\bottomrule
\end{tabular}
\caption{Macro~F1 for all 32 model--prompt configurations on Subtask~2 and Subtask~1 on the official test set (N\,=\,308). The overall best result is in \textbf{bold}; the best per-group result is \underline{underlined}.}
\label{tab:overview-full}
\end{table*}

\subsection{Qwen3 Thinking-Mode Diagnostic}
\label{sec:appendix-thinking}

To probe whether disabling native Qwen thinking mode made the main PEFT comparison overly pessimistic, we trained a dedicated Qwen3-32B checkpoint under the baseline setup and compared inference with \texttt{enable\_thinking=False} versus \texttt{enable\_thinking=True}. Because thinking-mode inference was approximately 4--5$\times$ slower (about 100 minutes for 100 samples in our setup) and the follow-up used different hardware (RTX PRO 6000 Blackwell rather than H100), we evaluated only stratified samples of 100 instances for the thinking runs. This diagnostic should therefore be read as within-checkpoint evidence about output stability and reasoning support rather than as a direct replacement for the main benchmark.

\begin{table}[t]
\centering
\small
\setlength{\tabcolsep}{4pt}
\begin{tabular}{l l c c c}
\toprule
\textbf{Split} & \textbf{Mode} & \textbf{N} & \textbf{Evasion F1} & \textbf{Clarity F1} \\
\midrule
Official Test & Non-thinking & 308 & 0.0734 & 0.2459 \\
Official Test & Thinking     & 100 & \textbf{0.2362} & \textbf{0.5127} \\
Blind Set     & Thinking     & 100 & \textbf{0.3700} & \textbf{0.5913} \\
\bottomrule
\end{tabular}
\caption{Diagnostic thinking-mode follow-up for Qwen3-32B. Non-thinking is evaluated on the full official test set; thinking is evaluated on stratified samples due to compute constraints, so direct cross-N comparisons should be interpreted cautiously.}
\label{tab:thinking-followup}
\end{table}

The non-thinking row in Table~\ref{tab:thinking-followup} refers to this diagnostic checkpoint under CoT JSON output format and is therefore distinct from the best Qwen3-32B configuration in the main paper. Its poor score is driven by output-format failure: 37.99\% of responses trigger a hard fallback to \textit{Explicit}. Enabling thinking reduces this to 0.00\% on the official-test sample and 1.00\% on the blind set (Table~\ref{tab:fallback}), indicating that native reasoning materially stabilizes output formatting as well as downstream classification.

\section{Prompt Templates}
\label{sec:appendix-prompts}

We use six distinct prompt templates across our experiments. The templates differ along two dimensions: (1)~output format (standard text vs.\ structured CoT JSON) and (2)~taxonomy presentation (hierarchical vs.\ flat). Below, we reproduce the exact text of each template. Placeholders \texttt{\{interview\_question\}}, \texttt{\{question\}}, and \texttt{\{answer\}} are filled at inference time.

\subsection{Standard Prompts (Fine-Tuned Models)}
\label{sec:prompt-standard}

\noindent\textbf{Zero-shot Standard.}
This prompt is used for fine-tuned models with standard output format (\texttt{Label: <label>}).

\begin{small}
\begin{verbatim}
You are a political discourse analyst. Your task
is to classify the evasion technique used in
Question-Answer pairs.

## Taxonomy (9 Evasion Labels)
Selected the best fitting label from the list
below:

1. **Explicit**: The respondent directly answers
   the specific question asked with the expected
   information.
2. **Implicit**: The answer is provided but
   requires inference (not stated plainly).
3. **General**: The respondent talks about the
   general topic but offers broad, vague
   platitudes instead of specific details.
4. **Dodging**: The respondent completely ignores
   the question content and shifts to an
   unrelated topic.
5. **Deflection**: The respondent acknowledges
   the topic but shifts focus to a specific
   tangent (e.g., attacking, self-praise) to
   avoid the core inquiry.
6. **Partial/half-answer**: The respondent
   addresses only one part of a multi-part
   question or a sub-issue, ignoring the main
   point.
7. **Declining to answer**: Explicit refusal to
   answer or stating they cannot comment.
8. **Claims ignorance**: The respondent states
   they do not know the answer or lack
   information.
9. **Clarification**: The respondent answers with
   a question to clarify context or meaning.

## Classification Guidelines
1. Check if answer directly addresses what
   question requests.
2. Explicit = direct answer. All others are
   evasion forms.
3. Distinguish: Implicit (implied) vs General
   (vague) vs Dodging (unrelated) vs Deflection
   (shifted focus).

## Output Format
Return exactly:
Label: <label>

## Analysis Task

Full Interview Context: {interview_question}

Specific Question: {question}

Answer: {answer}

Classify the evasion technique.
\end{verbatim}
\end{small}

\noindent\textbf{Few-shot Standard.}
Extends the zero-shot template by inserting six exemplars (one per representative label) between the taxonomy and the analysis task. Each exemplar contains a question, answer, and gold label. Due to space constraints, we show one representative exemplar below; the full set covers \textit{Explicit}, \textit{Deflection}, \textit{Partial/half-answer}, \textit{General}, \textit{Implicit}, and \textit{Claims ignorance}.

\begin{small}
\begin{verbatim}
### Example 1: Explicit
Question: How would you respond to accusations
  of containing China while pushing for
  diplomatic talks?
Answer: I don't want to contain China. I just
  want to make sure we have a relationship that
  is on the up and up.
Label: Explicit
\end{verbatim}
\end{small}

\subsection{CoT Prompts with Hierarchical Taxonomy (API Models)}
\label{sec:prompt-cot}

\noindent\textbf{Zero-shot CoT (Hierarchical).}
Used for API models. Includes a structured taxonomy grouped by clarity tier and requires JSON output with reasoning.

\begin{small}
\begin{verbatim}
You are an expert political discourse analyst
specializing in detecting equivocation and
evasion strategies in high-stakes interviews.
Your objective is to analyze the relationship
between a specific question and the respondent's
answer to classify the evasion technique used.

## The Taxonomy of Evasion
   (Classify into exactly ONE label)

### 1. Clear Reply
* **Explicit**: The respondent directly answers
  the specific question asked with the expected
  information.

### 2. Ambivalent Reply (Evasive or Vague)
* **Implicit**: The answer is provided but
  requires inference.
* **General**: The respondent talks about the
  general topic but offers broad, vague
  platitudes.
* **Partial/half-answer**: The respondent answers
  only one part of a multi-part question.
* **Dodging**: The respondent completely ignores
  the question content.
* **Deflection**: The respondent acknowledges the
  topic but pivots to a specific tangent.

### 3. Clear Non-Reply
* **Declining to answer**: Explicit refusal to
  answer.
* **Claims ignorance**: States they do not know.
* **Clarification**: Asks for clarification.

## Analysis Rules
1. **Directness Check**: Does the answer
   linguistically satisfy the interrogative
   specific of the question?
2. **Topic Fidelity**: Does the answer stay on
   the specific sub-topic of the question, or
   does it drift to a general theme?
3. **Reasoning First**: You must write out your
   analysis *before* deciding the label.

## Output Format
You must return a valid JSON object. Do not
include markdown formatting (like ```json).
Structure:
{
  "id": "string",
  "reasoning": "1. Directness Analysis: ...
    2. Topic Fidelity Analysis: ...
    3. Conclusion: ...",
  "evasion_label": "Exact Label String"
}

## YOUR TASK
Please analyze this specific QA pair:

[Full Context]: {interview_question}
(Note: Use this to understand the broader topic)

[Target Question]: {question}
(Note: Analyze the answer strictly against THIS
specific question)

[Respondent's Answer]: {answer}

Perform the analysis and return the JSON object.
\end{verbatim}
\end{small}

\noindent\textbf{Few-shot CoT (Hierarchical).}
Extends the zero-shot CoT template by inserting nine exemplars (one per label) with full reasoning traces. Each exemplar includes the target question, answer, and a complete JSON output containing directness analysis, topic fidelity analysis, and conclusion. Due to space, we show one exemplar:

\begin{small}
\begin{verbatim}
### Example 5: Dodging (Complete Topic Switch)
[Target Question]: Are you worried about the
  meeting between President Putin and Kim Jong
  Un, if that could mean Russia has more gains
  in the war in Ukraine?
[Respondent's Answer]: Look, I think China has a
  difficult economic problem right now ...
  we're not looking to hurt China, sincerely.
[JSON Output]:
{
  "id": "ex_5_dodging",
  "reasoning": "1. Directness Analysis: The
    question is about Putin, Kim Jong Un, and
    Ukraine. 2. Topic Fidelity Analysis: The
    answer talks entirely about China, Taiwan.
    Zero connection to Russia or North Korea.
    3. Conclusion: Complete topic switch.",
  "evasion_label": "Dodging"
}
\end{verbatim}
\end{small}

\subsection{CoT Prompts with Flat Taxonomy (API Models)}
\label{sec:prompt-flat}

\noindent\textbf{Zero-shot CoT (Flat / No Taxonomy).}
Identical structure to the hierarchical variant, but presents labels as a flat numbered list without grouping under clarity tiers. The key difference is the taxonomy section:

\begin{small}
\begin{verbatim}
## Taxonomy (9 Evasion Labels)
Selected the best fitting label from the list
below:

1. **Explicit**: The respondent directly answers
   the specific question asked with the expected
   information.
2. **Implicit**: The answer is provided but
   requires inference (not stated plainly).
3. **General**: The respondent talks about the
   general topic but offers broad, vague
   platitudes instead of specific details.
4. **Dodging**: The respondent completely ignores
   the question content and shifts to an
   unrelated topic.
5. **Deflection**: The respondent acknowledges
   the topic but shifts focus to a specific
   tangent to avoid the core inquiry.
6. **Partial/half-answer**: The respondent
   addresses only one part of a multi-part
   question or a sub-issue, ignoring the main
   point.
7. **Declining to answer**: Explicit refusal to
   answer or stating they cannot comment.
8. **Claims ignorance**: The respondent states
   they do not know the answer or lack
   information.
9. **Clarification**: The respondent answers with
   a question to clarify context or meaning.
\end{verbatim}
\end{small}

\noindent The analysis rules, output format, and few-shot exemplars remain identical to the hierarchical version. Only the taxonomy presentation differs.

\noindent\textbf{Few-shot CoT (Flat).} Combines the flat taxonomy above with the same nine exemplars and reasoning traces as the hierarchical few-shot variant.

\section{Computational Resources}
\label{sec:appendix-compute}

All fine-tuning experiments were conducted on a single NVIDIA H100 GPU (80\,GB VRAM). Using the Unsloth framework, parameter-efficient fine-tuning was exceptionally fast, with the maximum training time for the largest model (32B) taking under 18 minutes. However, the inference process was significantly slower due to the high volume of generation required for the reasoning outputs, with the few-shot chain-of-thought (CoT) prompting configuration being the most time-consuming to evaluate.

\paragraph{API Experiments.}
API experiments were executed via asynchronous batch endpoints with no local GPU requirements. For Grok-4-Fast, requests were submitted through the xAI Batch API in chunks of up to 25{,}000 items; each chunk typically completed within 15 minutes, with a worst-case observed latency of under 30 minutes per batch. For DeepSeek-V3.2, requests were submitted via the DeepSeek batch API with similarly low latency.

\paragraph{API Cost.}
We compute per-sample costs from the \texttt{cost\_in\_usd\_ticks} field recorded in all Grok batch result files, calibrated against a known reference sample (\$0.0006 for one reasoning request). Table~\ref{tab:api-cost} reports the actual cost for all 8 Grok configurations (N\,=\,308 each). Reasoning-mode requests average \$0.000411 per sample versus \$0.000178 for non-reasoning, a ratio of approximately 2.3$\times$. The best-performing configuration (Grok-4-Fast Reasoning, Few-shot CoT) costs \$0.1359 for the full test set. The total expenditure for all 8 Grok configurations is \$0.70. DeepSeek-V3.2 API costs are of a similar order of magnitude, giving a total API expenditure across all 16 configurations of approximately \$1.50.

\begin{table}[h]
\centering
\small
{\setlength{\tabcolsep}{4pt}
\begin{tabular}{l l c c}
\toprule
\textbf{Config} & \textbf{Mode} & \textbf{Total (\$)} & \textbf{Per sample (\$)} \\
\midrule
Zero-shot flat  & Non-reasoning & 0.0417 & 0.000135 \\
Zero-shot hier. & Non-reasoning & 0.0461 & 0.000150 \\
Few-shot flat   & Non-reasoning & 0.0647 & 0.000210 \\
Few-shot hier.  & Non-reasoning & 0.0672 & 0.000218 \\
\midrule
Zero-shot flat  & Reasoning     & 0.1148 & 0.000373 \\
Zero-shot hier. & Reasoning     & 0.1158 & 0.000376 \\
Few-shot flat   & Reasoning     & 0.1175 & 0.000455 \\
Few-shot hier.  & Reasoning     & 0.1359 & 0.000441 \\
\midrule
\textbf{Total} & & \textbf{0.7036} & \\
\bottomrule
\end{tabular}
}
\caption{Actual API cost for all 8 Grok-4-Fast configurations on the official test set (N\,=\,308), derived from \texttt{cost\_in\_usd\_ticks} in batch result files.}
\label{tab:api-cost}
\end{table}

\section{Confusion Matrix and Qualitative Error Analysis}
\label{sec:appendix-cm}

\begin{figure}[h]
\centering
\includegraphics[width=\linewidth]{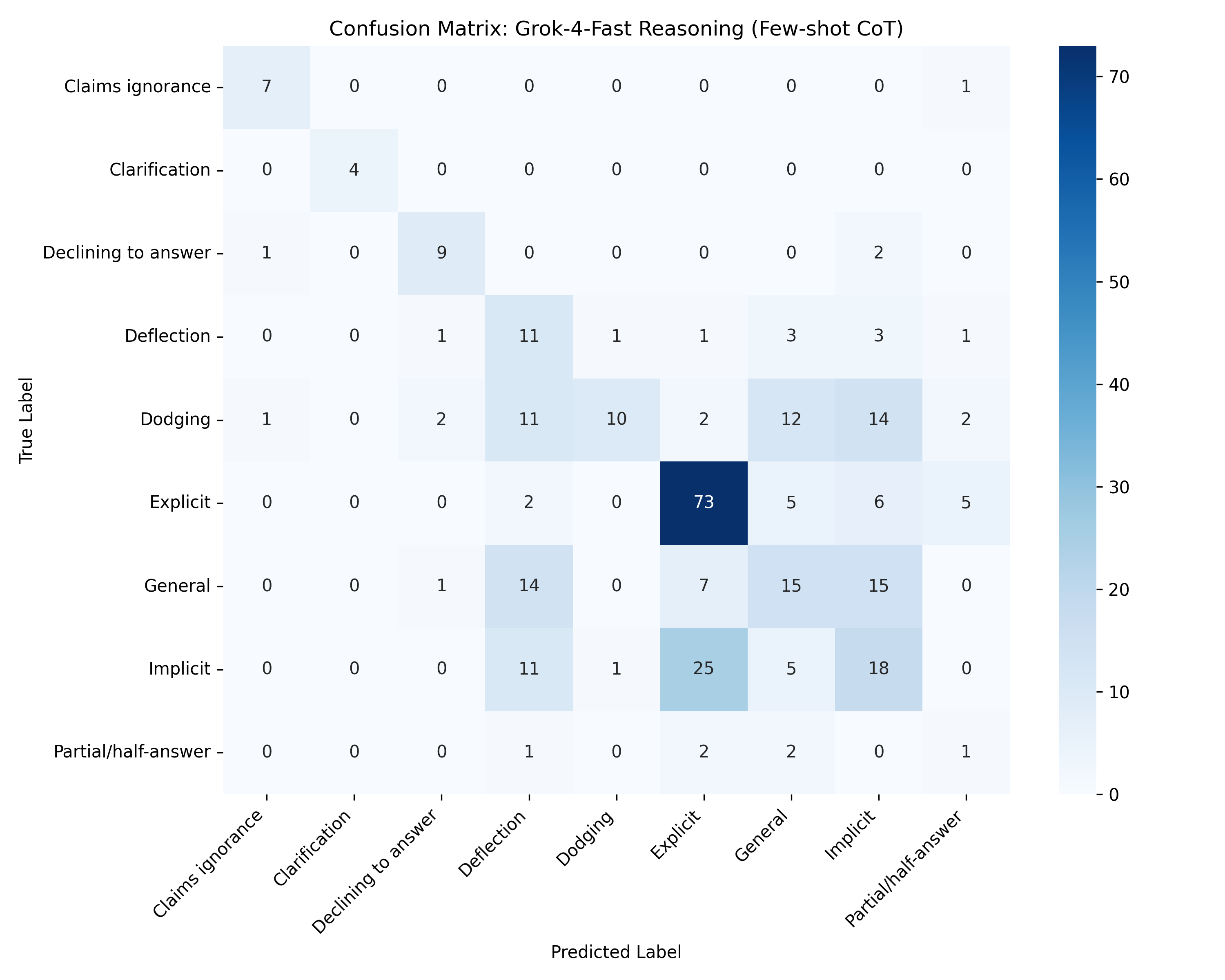}
\caption{Confusion matrix of Subtask 2 predictions versus gold labels.}
\label{fig:cm}
\end{figure}

Figure~\ref{fig:cm} presents the full confusion matrix for Subtask 2 using our best configuration (Grok-4-Fast Reasoning, Few-shot CoT) on the official test set (N\,=\,308).

\subsection{Case Study Discussion: Reasoning Traces vs.\ Human Pragmatics}

To understand why the best-performing models still struggle with ambivalent evasion categories, we analyze the explicit reasoning traces generated by Grok-4-Fast. Tables~\ref{tab:qual1} and~\ref{tab:qual2} present two case studies where the model's logical deduction cleanly misaligned with human pragmatic judgment.

\subsubsection{Case Study 1: Dodging vs. Deflection Confusion}

The first common failure mode involves distinguishing between \textit{Dodging} (complete topic switch) and \textit{Deflection} (acknowledged topic with tangential pivot). Table~\ref{tab:qual1} illustrates this challenge. The model successfully detects the pivot to the PATRIOT Act policy discussion. However, it fails to weight the \textit{severity} of the pivot correctly. Because the politician mentions "difference of attitude" and "meant nothing personal", the model assumes the original topic (the apology demand) was sufficiently acknowledged, thus assigning \textit{Deflection}. 

Human annotators, conversely, recognized that the ensuing 200-word policy defense entirely \textit{replaced} the apology demand rather than merely deflecting from it, constituting a complete \textit{Dodge}. This reveals a fundamental limitation: the model's decompositional reasoning breaks down pragmatic judgment into discrete logical steps, losing the holistic assessment of response adequacy that humans perform naturally. The mere mention of the original topic's semantic field (political disagreement) is insufficient to constitute topic acknowledgment if the core question (about a specific apology) is never addressed.

\begin{table}[!ht]
\centering
\footnotesize
{\setlength{\tabcolsep}{3pt}
\begin{tabular}{p{0.95\columnwidth}}
\toprule
\textbf{Case Study 1: Dodging vs. Deflection} \\
\midrule
\textbf{Q:} Do you think the Republican leader in the House owes somebody an apology? \\
\addlinespace[0.5em]
\textbf{A:} I wouldn't have exactly put it that way. But I do believe there's a difference of attitude. I mean, take the PATRIOT Act, for example [...] I'm confident the leader, you know, meant nothing personal. \\
\addlinespace[0.5em]
\textbf{Gold:} \textit{Dodging} \quad \textbf{Pred:} \textit{Deflection} \\
\addlinespace[0.5em]
\textbf{Reasoning:} \textit{1. Directness:} The question is yes/no about whether the leader owes an apology. The respondent does not explicitly affirm or deny. \textit{2. Topic Fidelity:} Acknowledges controversy but pivots to PATRIOT Act, shifting focus to broader policy differences. \textit{3. Conclusion:} Acknowledges topic but deflects via policy tangent. \\
\bottomrule
\end{tabular}
}
\caption{Model misclassifies complete topic replacement (\textit{Dodging}) as acknowledged pivot (\textit{Deflection}).}
\label{tab:qual1}
\end{table}

\subsubsection{Case Study 2: General vs. Implicit Confusion}

The second failure mode involves the boundary between \textit{General} (vague platitudes) and \textit{Implicit} (answer requiring inference). Table~\ref{tab:qual2} demonstrates this challenge. The model logically extracts positive sentiment from the respondent's mention of "priority lists" and "alliances", treating these cues as a solvable puzzle that yields an \textit{Implicit} answer (i.e., "No, I am not concerned because the region is a priority").

However, human annotators interpreted these same statements as political platitudes that offered no concrete answer regarding actual fiscal pressures, labeling it \textit{General}. The critical distinction lies in \textit{specificity}: an implicit answer still addresses the question's core inquiry through contextual cues, whereas a general answer substitutes broad rhetoric for substantive response. The model's step-by-step deduction over-extracts specific intent from diffuse political language, failing to recognize when semantic content is merely decorative rather than informative.

\textbf{Implications for Future Work:}
These case studies reveal that chain-of-thought reasoning, while improving overall accuracy, introduces systematic biases toward over-interpretation. The model's explicit logical decomposition paradoxically undermines its ability to recognize absence of substance—a skill that requires holistic pragmatic judgment rather than compositional semantic analysis. Future approaches might benefit from incorporating adversarial training examples that specifically target these boundary cases, or developing hybrid architectures that combine logical decomposition with gestalt-level assessment of response adequacy.

\begin{table}[!ht]
\centering
\footnotesize
{\setlength{\tabcolsep}{3pt}
\begin{tabular}{p{0.95\columnwidth}}
\toprule
\textbf{Case Study 2: General vs. Implicit} \\
\midrule
\textbf{Q:} Are you concerned that fiscal pressures may compromise US plans for the region? \\
\addlinespace[0.5em]
\textbf{A:} With respect to Europe, I'm deeply concerned [...] with respect to our budget, there's a reason why I'm spending this time out here in Asia [...] President Obama being here shows he values the US role in this region and our alliance [...]. \\
\addlinespace[0.5em]
\textbf{Gold:} \textit{General} \quad \textbf{Pred:} \textit{Implicit} \\
\addlinespace[0.5em]
\textbf{Reasoning:} \textit{1. Directness:} The question asks yes/no about fiscal concerns. Respondent doesn't state concern but provides reassurance by emphasizing regional priority. \textit{2. Topic Fidelity:} Transitions to budget and Asia-Pacific commitment, staying relevant. \textit{3. Conclusion:} Answer implies lack of concern through priority statements—requires inference, fitting \textit{Implicit}. \\
\bottomrule
\end{tabular}
}
\caption{Model over-extracts specific intent from platitudes, misclassifying \textit{General} as \textit{Implicit}.}
\label{tab:qual2}
\end{table}

\subsubsection{Case Study 3: Model Prediction Aligns with Minority Annotator}
\label{sec:appendix-cs3}

The preceding case studies illustrate model failures. However, annotator disagreement also creates cases where a model's prediction, while diverging from the gold label, is arguably more defensible than the majority annotation. We present one such instance to contextualize the inherent subjectivity of the task.

Table~\ref{tab:qual3} shows a QA pair where two of three annotators labeled the response \textit{Dodging}, establishing the gold label, while the third annotator labeled it \textit{Implicit}. The model (Grok-4-Fast Reasoning, Few-shot CoT) predicted \textit{Implicit}, aligning with the minority annotator.

The response ``I'm going to stay out of Connecticut'' does not explicitly answer the yes/no question about campaigning. However, it is \emph{topically anchored}: Connecticut is Senator Lieberman's home state and the location where any such campaign would take place. The geographic reference serves as an indirect indexical that conveys a clear position (non-involvement) through inference—the defining criterion of \textit{Implicit}. By contrast, \textit{Dodging} requires a \emph{complete} topic switch with no connection to the original question; here the connection is direct, even if compressed.

The majority annotation of \textit{Dodging} likely reflects annotators prioritizing the absence of an explicit yes/no commitment and the brevity of the response, rather than the semantic content of the geographic anchor. This case illustrates that some majority-vote labels are not uniquely determined under the taxonomy, and that model errors computed against a single gold label may include pragmatically defensible alternatives.

\begin{table}[H]
\centering
\footnotesize
{\setlength{\tabcolsep}{3pt}
\begin{tabular}{p{0.95\columnwidth}}
\toprule
\textbf{Case Study 3: Annotator Disagreement — \textit{Implicit} vs.\ \textit{Dodging}} \\
\midrule
\textbf{Q:} Would you campaign against Senator Joe Lieberman on Iraq? \\
\addlinespace[0.5em]
\textbf{A:} I'm going to stay out of Connecticut. \\
\addlinespace[0.5em]
\textbf{Gold:} \textit{Dodging} \quad \textbf{Pred:} \textit{Implicit} \quad \textbf{Annotators:} \textit{Dodging} / \textit{Dodging} / \textit{Implicit} \\
\addlinespace[0.5em]
\textbf{Model Reasoning:} \textit{1. Directness:} The question asks yes/no about campaigning. The answer does not explicitly state yes or no. \textit{2. Topic Fidelity:} The response references Connecticut, Lieberman's home state and the site of any such campaign, directly tying to the question context. \textit{3. Conclusion:} Staying out of Connecticut implies non-involvement in campaigns there—an answer conveyed through geographic inference, fitting \textit{Implicit}. \\
\bottomrule
\end{tabular}
}
\caption{A case where the model's \textit{Implicit} prediction aligns with the minority annotator. The geographic anchor (Connecticut = Lieberman's state) constitutes topical relevance, making \textit{Dodging} (complete topic switch) a debatable majority label under the taxonomy definition.}
\label{tab:qual3}
\end{table}
\clearpage

%% file: custom.bib
@misc{yang2025qwen3technicalreport,
  title={Qwen3 Technical Report},
  author={Yang, An and Li, Anfeng and Yang, Baosong and Zhang, Beichen and Hui, Binyuan and Zheng, Bo and Yu, Bowen and Gao, Chang and Huang, Chengen and Lv, Chenxu and Zheng, Chujie and Liu, Dayiheng and Zhou, Fan and Huang, Fei and Hu, Feng and Ge, Hao and Wei, Haoran and Lin, Huan and Tang, Jialong and Yang, Jian and Tu, Jianhong and Zhang, Jianwei and Yang, Jianxin and Yang, Jiaxi and Zhou, Jing and Zhou, Jingren and Lin, Junyang and Dang, Kai and Bao, Keqin and Yang, Kexin and Yu, Le and Deng, Lianghao and Li, Mei and Xue, Mingfeng and Li, Mingze and Zhang, Pei and Wang, Peng and Zhu, Qin and Men, Rui and Gao, Ruize and Liu, Shixuan and Luo, Shuang and Li, Tianhao and Tang, Tianyi and Yin, Wenbiao and Ren, Xingzhang and Wang, Xinyu and Zhang, Xinyu and Ren, Xuancheng and Fan, Yang and Su, Yang and Zhang, Yichang and Zhang, Yinger and Wan, Yu and Liu, Yuqiong and Wang, Zekun and Cui, Zeyu and Zhang, Zhenru and Zhou, Zhipeng and Qiu, Zihan},
  year={2025},
  eprint={2505.09388},
  archivePrefix={arXiv},
  primaryClass={cs.CL},
  url={https://arxiv.org/abs/2505.09388}
}

@inproceedings{ferracane2021didanswer,
  title        = {Did They Answer? Subjective Acts and Intents in Conversational Discourse},
  author       = {Ferracane, Elisa and Durrett, Greg and Li, Junyi Jessy and Erk, Katrin},
  booktitle    = {Proceedings of the 2021 Conference of the North American Chapter of the Association for Computational Linguistics: Human Language Technologies},
  year         = {2021},
  pages        = {1272--1286},
  publisher    = {Association for Computational Linguistics},
  address      = {Online},
  doi          = {10.18653/v1/2021.naacl-main.129},
  url          = {https://aclanthology.org/2021.naacl-main.129}
}

@article{bull1993hownotanswer,
  title        = {How Not to Answer Questions in Political Interviews},
  author       = {Bull, Peter and Mayer, Kate},
  journal      = {Political Psychology},
  volume       = {14},
  number       = {4},
  pages        = {651--666},
  year         = {1993},
  publisher    = {Wiley},
  doi          = {10.2307/3791379},
  url          = {https://www.jstor.org/stable/3791379}
}

@article{rasiah2010frameworkevasion,
  title        = {A Framework for the Systematic Analysis of Evasion in Parliamentary Discourse},
  author       = {Rasiah, Parameswary},
  journal      = {Journal of Pragmatics},
  volume       = {42},
  number       = {3},
  pages        = {664--680},
  year         = {2010},
  publisher    = {Elsevier},
  doi          = {10.1016/j.pragma.2009.07.010},
  url          = {https://doi.org/10.1016/j.pragma.2009.07.010}
}

@book{bull2003microanalysis,
  title={The Microanalysis of Political Communication: Claptrap and Ambiguity},
  author={Bull, Peter},
  publisher={Routledge},
  year={2003}
}

@article{clayman2001answers,
  title={Answers and evasions},
  author={Clayman, Steven E.},
  journal={Language in Society},
  volume={30},
  number={3},
  pages={403--442},
  year={2001}
}

@inproceedings{thomas-etal-2024-never,
  title     = {``{I} Never Said That'': A dataset, taxonomy and baselines on response clarity classification},
  author    = {Thomas, Konstantinos and Filandrianos, Giorgos and Lymperaiou, Maria and Zerva, Chrysoula and Stamou, Giorgos},
  editor    = {Al-Onaizan, Yaser and Bansal, Mohit and Chen, Yun-Nung},
  booktitle = {Findings of the Association for Computational Linguistics: EMNLP 2024},
  month     = nov,
  year      = {2024},
  address   = {Miami, Florida, USA},
  publisher = {Association for Computational Linguistics},
  url       = {https://aclanthology.org/2024.findings-emnlp.300/},
  doi       = {10.18653/v1/2024.findings-emnlp.300},
  pages     = {5204--5233},
}

@misc{thomas2026semeval2026task6clarity,
  title         = {{SemEval-2026 Task 6: CLARITY -- Unmasking Political Question Evasions}},
  author        = {Thomas, Konstantinos and Filandrianos, Giorgos and Lymperaiou, Maria and Zerva, Chrysoula and Stamou, Giorgos},
  year          = {2026},
  eprint        = {2603.14027},
  archivePrefix = {arXiv},
  primaryClass  = {cs.CL},
  url           = {https://arxiv.org/abs/2603.14027},
}

@inproceedings{dayanik2022improving,
  title={Improving Neural Political Statement Classification with Class Hierarchical Information},
  author={Dayanik, Erenay and Blessing, Andr{\'e} and Blokker, Nico and Haunss, Sebastian and Kuhn, Jonas},
  booktitle={Findings of ACL},
  year={2022},
  doi={10.18653/v1/2022.findings-acl.186}
}

@inproceedings{hu2022lora,
  title={LoRA: Low-Rank Adaptation of Large Language Models},
  author={Hu, Edward J. and Shen, Yelong and Wallis, Phillip and Allen-Zhu, Zeyuan and Li, Yuanzhi and Wang, Shean and Wang, Lu and Chen, Weizhu},
  booktitle={International Conference on Learning Representations (ICLR)},
  year={2022}
}

@inproceedings{dettmers2023qlora,
  title={QLoRA: Efficient Finetuning of Quantized LLMs},
  author={Dettmers, Tim and Pagnoni, Artidoro and Holtzman, Ari and Zettlemoyer, Luke},
  booktitle={Advances in Neural Information Processing Systems},
  volume={36},
  year={2023}
}

@inproceedings{wei2022chain,
  title={Chain-of-Thought Prompting Elicits Reasoning in Large Language Models},
  author={Wei, Jason and Wang, Xuezhi and Schuurmans, Dale and Bosma, Maarten and Ichter, Brian and Xia, Fei and Chi, Ed H. and Le, Quoc V. and Zhou, Denny},
  booktitle={NeurIPS},
  year={2022},
  doi={10.48550/arxiv.2201.11903}
}

@inproceedings{wang2023plan,
  title={Plan-and-Solve Prompting: Improving Zero-Shot Chain-of-Thought Reasoning by Large Language Models},
  author={Wang, Lei and Xu, Wanyu and Lan, Yihuai and Hu, Zhiqiang and Lan, Yunshi and Lee, Roy Ka-Wei and Lim, Ee-Peng},
  booktitle={ACL},
  year={2023},
  doi={10.18653/v1/2023.acl-long.147}
}

@inproceedings{li2025reinforcement,
  title={A Reinforcement Learning Framework for Cross-Lingual Stance Detection Using Chain-of-Thought Alignment},
  author={Li, Binghui and Zou, Minghui and Zhang, Xiaowang and Chen, Shizhan and Feng, Zhiyong},
  booktitle={Findings of ACL},
  year={2025},
  doi={10.18653/v1/2025.findings-acl.1115}
}

@misc{han2024unsloth,
  title={Unsloth: Fast and Memory-Efficient LLM Fine-tuning},
  author={Han, Daniel and Han, Michael},
  year={2024},
  url={https://github.com/unslothai/unsloth}
}

@inproceedings{yeh2000more,
  title     = {More Accurate Tests for the Statistical Significance of Result Differences},
  author    = {Yeh, Alexander},
  booktitle = {Proceedings of the 18th International Conference on Computational Linguistics (COLING)},
  year      = {2000},
  pages     = {947--953},
  publisher = {Association for Computational Linguistics},
  url       = {https://aclanthology.org/C00-2137/},
}

@article{mcnemar1947note,
  title   = {Note on the Sampling Error of the Difference between Correlated Proportions or Percentages},
  author  = {McNemar, Quinn},
  journal = {Psychometrika},
  volume  = {12},
  number  = {2},
  pages   = {153--157},
  year    = {1947},
  doi     = {10.1007/BF02295996},
}
